%% file: main.tex
\documentclass{article}

\usepackage[final]{neurips_2024}
\usepackage{hyperref}       % hyperlinks
\usepackage{graphicx}
\usepackage{amsmath}
\usepackage{multirow}

\usepackage[utf8]{inputenc} % allow utf-8 input
\usepackage[T1]{fontenc}    % use 8-bit T1 fonts

\usepackage{url}            % simple URL typesetting
\usepackage{booktabs}       % professional-quality tables
\usepackage{amsfonts}       % blackboard math symbols
\usepackage{nicefrac}       % compact symbols for 1/2, etc.
\usepackage{microtype}      % microtypography
\usepackage{xcolor}         % colors

\title{Leveraging Vision-Language Foundation Models to Reveal Hidden Image-Attribute Relationships in Medical Imaging}

\author{%
  Amar ~Kumar$^*$\\
  McGill University\\
  MILA-Quebec AI Institute\\
  \texttt{amar.kumar@mail.mcgill.ca} \\
  \And
  Anita ~Kriz$^*$\\
  McGill University\\
  MILA-Quebec AI Institute\\
  \texttt{anita.kriz@mail.mcgill.ca} \\
  \AND
  Barak ~Pertzov\\
  McMaster University\\
  \texttt{pertzovb@mcmaster.ca} 
  \And
  Tal ~Arbel\\
  McGill University\\
  MILA-Quebec AI Institute\\
  \texttt{tal.arbel@mcgill.ca} \\
}

\begin{document}

\maketitle
\renewcommand{\thefootnote}{\fnsymbol{footnote}}
\footnotetext[1]{Equal contribution.}
\begin{abstract}
Vision-language foundation models (VLMs) have shown impressive performance in guiding image generation through text, with emerging applications in medical imaging. In this work, we are the first to investigate the question: \textit{`Can fine-tuned foundation models help identify critical, and possibly unknown, data properties?'} By evaluating our proposed method on a chest x-ray dataset, we show that these models can generate high-resolution, precisely edited images compared to methods that rely on Structural Causal Models (SCMs) according to numerous metrics. For the first time, we demonstrate that fine-tuned VLMs can reveal hidden data relationships that were previously obscured due to available metadata granularity and model capacity limitations. Our experiments demonstrate both the potential of these models to reveal underlying dataset properties while also exposing the limitations of fine-tuned VLMs for accurate image editing and susceptibility to biases and spurious correlations. 
\end{abstract}

\input{sec/1_intro}
\input{sec/2_methodology}

\input{sec/3_experiments}

%%%%%%%%%%%%%%%%%%%%%%%%%%%%%%%%%%%%%%%%%%%%%%%%%%%%%%%%%%%%
\newpage
\begin{ack}
The authors are grateful for funding provided by the Natural Sciences and Engineering Research Council of Canada, the Canadian Institute for Advanced Research (CIFAR) Artificial Intelligence Chairs program, Mila - Quebec AI Institute, Google Research, Calcul Qu\'ebec, Fonds de recherche du Qu\'ebec (FRQNT),  and the Digital Research Alliance of Canada.
\end{ack}

{
    \bibliographystyle{ieeenat_fullname}
    \bibliography{main}
}

%\appendix

%\section{Appendix / supplemental material}

%Optionally include supplemental material (complete proofs, additional experiments and plots) in appendix.
%All such materials \textbf{SHOULD be included in the main submission.}

%%%%%%%%%%%%%%%%%%%%%%%%%%%%%%%%%%%%%%%%%%%%%%%%%%%%%%%%%%%%

\end{document}

%% file: sec/1_intro.tex
\section{Introduction}
\label{sec:intro}

% Understanding disease progression and predicting future patient outcomes is essential fo establishing treatment plans and enabling early intervention in medical imaging. Counterfactual image generation offers a promising approach to meet these objectives by enabling clinicians to explore "what-if" questions about a patient~\cite{ganguly2023review,castro2020causality}. For example, generating an X-ray for queries like \textit{"What will this patient's scan look like in 1 year"} could be used to  identity currently unknown disease markers ~\cite{castro2020causality,discoverysanchez2022,discoveryShen} and determine optimal treatment strategies~\cite{treatmentfeuerriegel2024,treatmentmsaouel2022}, ultimately transforming healthcare practices. However, distinguishing causation from correlation in medical data is inherently challenging. A model that relies solely on correlations in training data may inadvertently alter unrelated attributes -- such as the patient's sex -- due to spurious associations, resulting in inaccurate and unreliable outputs. In contrast, we expect a true counterfactual generative model to capture how biological ageing affects \textit{both} anatomical structures and disease progression, while remaining agnostic to unrelated attributes -- even when those attributes might be influenced by common confounders.
% \begin{quote}
%     \textit{A machine learning model is only as good as the data it is fed.} \\
%     \hspace*{\fill} \textsc{Reynold Xin}
% \end{quote}
Dataset biases present a significant challenge in the development of trustworthy machine learning models in healthcare applications. These biases can manifest as spurious correlations -- misleading patterns between attributes of the data -- which cause the model to learn incorrect associations that do not reflect true clinical relationships. 
% This can lead to propagating diagnostic errors and reduce clinical reliability. 
For instance, algorithms trained on chest X-ray learned to classify pneumothorax based on the presence of a chest drain,  which is inserted \textit{after} pneumothorax diagnosis to treat it, and thus does not reflect the true pathological disease markers \cite{pneumo_shortcut}. Similarly, dermatological diagnostic models demonstrated significantly decreased accuracy when analyzing underrepresented skin types, reducing diagnostic reliability \cite{Kleinberg2022RacialUI}. These learned biases can lead to potentially harmful clinical outcomes when models are deployed in real-world healthcare settings, where they will fail to generalize beyond the underlying distributions of their training data~\cite{yoon2023domain}. A particularly concerning aspect of these biases is that they may be unobserved within the metadata, making them difficult to identify through standard statistical analyses. By developing methodologies to identify and mitigate dataset bias, we can address the critical need for robust and reliable AI tools prior to their deployment in clinical settings. %TODO: do we need something here about how you may not even be able to see the spurious relations. TODO: add something about, that this can be somewhat discovered when you do subgroup analysis of your different labels, but impossible when you do not have access to the lables.

Causal-based approaches aim to answer causal queries and consequentially overcome challenges with spurious correlations by explicitly enforcing attribute relationships within a causal framework. These methods incorporate causal principles by embedding Structural Causal Models (SCMs) into the generation process \cite{pawlowski2020deep,glocker2023algorithmic,xia2021causal} and adhering to Pearl's Causal Hierarchy Theorem (CHT)\cite{pearl2000causality,pearl2009causality} to generate results. However, many SCM-based methods still require external classifiers to function effectively \cite{pawlowski2020deep, pan2024counterfactual}. Moreover, despite their theoretical rigor, the assumptions these methods require for theoretical validity often become unrealistic in real-world applications. In particular, their constrained ability to generate high-resolution images results in model outputs that may not capture underlying dataset biases due to limited editing precision. 

Foundation models, trained on vast and diverse datasets, have demonstrated remarkable generative capabilities in both computer vision~\cite{yuan2021florence} and medical imaging~\cite{azad2023foundational,xu2023elixr,liu2023radiology}. Two significant breakthroughs are particularly relevant: (i) enhanced image generation capabilities that enables the generation of unprecedented high-resolution images \cite{kumar2025prism}, and (ii) precise targeted editing of specific image attributes \cite{lang2021explaining}. Despite these capabilities, important questions remain about what (and how) these models learn (e.g. the field of mechanistic interpretability) and how the underlying data distribution impacts them. 

In this work, we explore the strengths and limitations of fine-tuned VLMs in generating high-resolution, realistic medical images while analyzing their dependence on, and ability to reveal, underlying dataset properties.  We first demonstrate that a non-SCM VLM model can generate high-resolution, realistic images faithful to the original (factual) images when compared to an SCM-based model. We quantitatively evaluate both methods using perceptual similarity to the factual image, pixel-wise differences that measure fidelity to expected transformations, and effectiveness ensuring counterfactuals induce the intended change while preserving unrelated attributes. Beyond demonstrating high-quality image editing with fine-tuned VLMs, we further analyze how dataset properties influence their learning process. By leveraging these high-capacity models for precise image editing, we:
(i) \textbf{Reveal Dataset Properties} that may not be explicitly captured in metadata labels, and thus not detectable through traditional statistical analyses, and (ii) \textbf{Demonstrate Failure Modes} that show how these models deviate from the true data-generating process, revealing their reliance on spurious correlations and dataset-specific artifacts~\cite{practicalcausal}.

% In this work, we compare our non-SCM VLM model to a SCM model, can generate high-resolution, realistic images faithful to the original (factual) image. We evaluate both methods using (1) \textbf{perceptual similarity} to the factual image, (2) \textbf{pixel-wise differences} that measure fidelity to expected transformations, and (3) \textbf{effectiveness} ensuring counterfactuals induce the intended change while preserving unrelated attributes.
% For the first time, through controlled image modifications, we reveal previously hidden data patterns and relationships that were previously obscured by the granularity of the metadata (and thus not detectable by conventional statistics) and the low capacity of earlier models. This in turn demon
% % While prior research has explored the extent to which foundation models learn causal relationships~\cite{mechanisticinterpretability} and whether they can answer causal queries~\cite{llmwillig2022foundationmodelstalkcausality,llmkhetan2021causalbertlanguage,llm2023causal}, recent studies suggest that Large Language Models (LLMs) struggle with causal reasoning~\cite{llmjin2024can,llmashwani2024causeeffectlargelanguage}.

\begin{figure}[h]
    \centering
    \includegraphics[width=\textwidth]{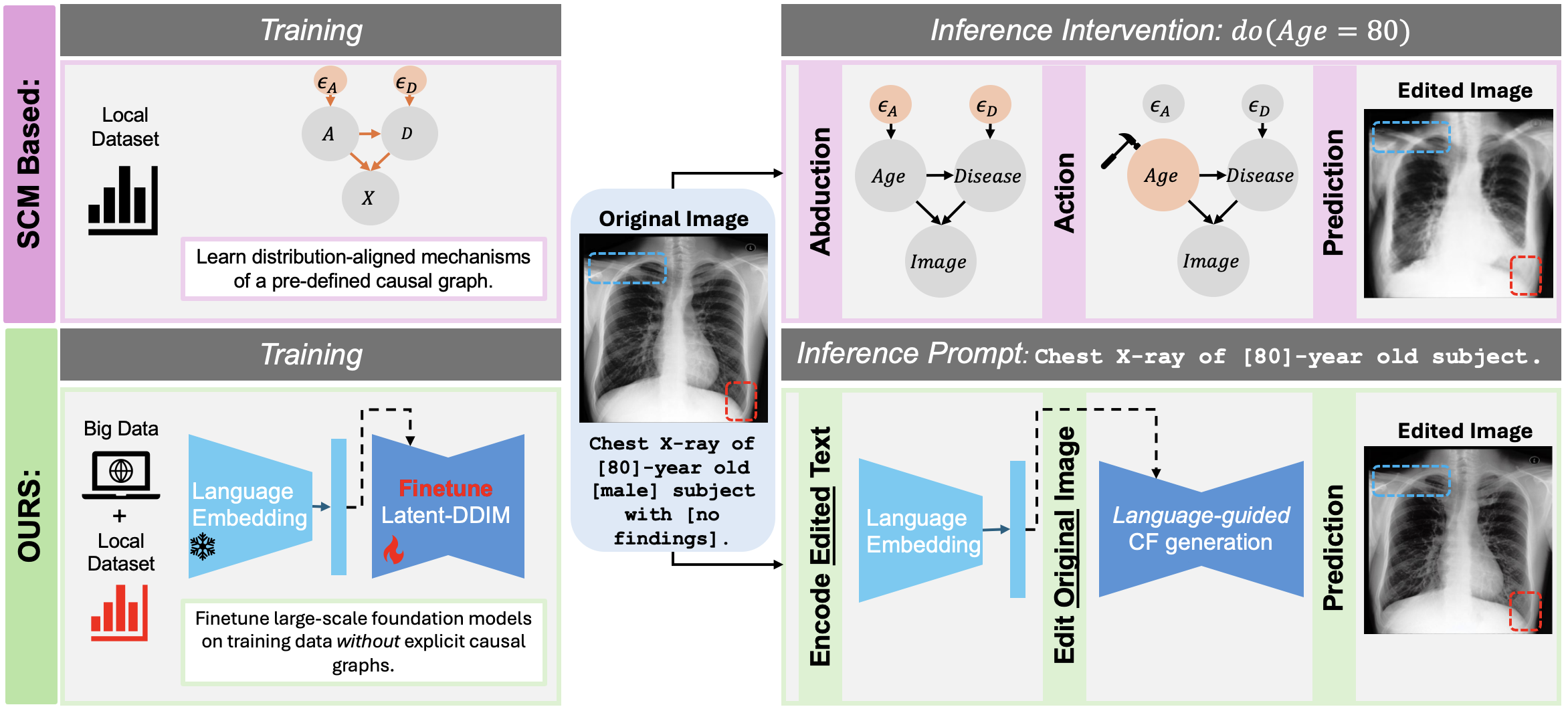}
    \caption{Comparison of counterfactual generation in the SOTA structural causal model (SCM) and a foundation vision-language model. Inference involves generating images with increased age, which may have downstream effects on disease. \textcolor{red}{Boxes} highlight changes related to disease state, while \textcolor[HTML]{639CFF}{boxes} indicate changes associated with age (e.g., decreased bone and tissue density). Given that the SCM model is built upon an HVAE and cannot generate high-resolution images, it is limited in its ability to capture plausible counterfactual generation at finer detail levels.}
    \label{fig:intro}
\end{figure}

%% file: sec/2_methodology.tex
\section{Methodology}
\label{sec:methodology}

\subsection{Background}
\noindent\textbf{Structural Causal Models \& Counterfactual Generation} Our work investigates how fine-tuned vision-language foundation models can generate high-resolution, faithful counterfactual images. We discuss counterfactual generation using Structural Causal Models (SCMs) to provide context. %Formally, an SCM is defined as a 4-tuple \( \mathcal{M} = (\mathbf{V}, \mathbf{U}, \mathbf{F}, \mathcal{G}) \), where \( \mathbf{V} \) represents a set of endogenous variables, \( \mathbf{U} \) denotes a set of exogenous (noise) variables, \( \mathbf{F} = {f_i} \) is a set of structural functions governing causal relationships, and \( \mathcal{G} \) is a directed acyclic graph (DAG) that encodes the dependencies among variables. Each endogenous variable \( V_i \) is determined by its causal parents \( \text{Pa}(V_i) \) and an exogenous noise term \( U_i \) through a structural equation of the form \( V_i = f_i(\text{Pa}(V_i), U_i) \).
These SCMs enable counterfactual reasoning for answering questions such as: \textit{What would image \( X \) look like if attribute \( A_i = a \) had instead been \( A_i = a' \)?} This is achieved by intervening on specific variables while preserving the underlying causal mechanisms. According to Pearl's Causal Hierarchy Theorem (CHT), counterfactual generation follows three key steps: (i) \textbf{Abduction:} The structural equations are inverted to infer the exogenous noise variables \( U \) that explain the observed image and attributes;(ii) \textbf{Action:} An intervention is performed on attribute \( A \), changing its value from \( A = a \) to \( A = a' \), represented using the do-operator as \( do(A = a') \); (iii) \textbf{Prediction:} The modified attribute \( A = a' \) is propagated through the structural equations, generating a new counterfactual image \( X^{a'} = f(A = a', U) \), while preserving \( U \) to maintain consistency with the original latent factors.

\noindent\textbf{Stable Diffusion} Stable Diffusion~\cite{Rombach_2022_CVPR}, our chosen generative model for counterfactual generation, is comprised of four components: (i) an \textit{Image encoder} that transforms input images into low-dimensional latent representations capturing essential features; (ii) a \textit{CLIP text encoder} that provides text conditioning to guide generation and ensure alignment between synthesized images and text prompts; (iii) a \textit{U-net denoiser} that forms the core of the reverse diffusion process, receiving latent representations and iteratively removing Gaussian noise across multiple timesteps by predicting and eliminating noise from the latent representation. In particular, Denoising Diffusion Implicit Models (DDIM) allows for a more efficient and deterministic sampling process by eliminating the need for all timesteps in the reverse diffusion process, resulting in faster generation with high-quality outputs; and (iv) an \textit{Image decoder} that converts denoised latents back to the original image space, producing high-resolution output images. By utilizing a Stable Diffusion model pre-trained on extensively large datasets, we can leverage its learned representations. Fine-tuning the model on our dataset then allows us to benefit from both the robustness of the pre-trained network and the adaptability to our specific task.

% We briefly describe Stable Diffusion~\cite{Rombach_2022_CVPR} , the generative model which we utilize for counterfactual generation. Stable Diffusion~\cite{Rombach_2022_CVPR} consists of four key components - (i) \textit{Image encoder} responsible for encoding input images into latent space (low dimensional) representation that captures essential features; (ii) \textit{Contrastive Language-Image Pre-training (CLIP) text encoder} provides the text conditioning to the Stable Diffusion model which in turn guides the generation process ensuring the synthesized images aligns with the text prompt. Note, that the CLIP model is already pre-trained on a large text corpus; (iii)\textit{U-net denoiser} is the core of the denoising process in reverse diffusion. The U-net receives the image's latent representation and successively denoises the Gaussian noise over multiple timesteps. At each timestep, the U-net predicts the noise that needs to be removed from the latent representation to refine and generate the output image's latent iteratively; and (iv) \textit{Image decoder} takes the denoised latent from the reverse diffusion step and maps the latent space back to the original image space, generating the final high-resolution image.

\subsection{Training \& Inference using Stable Diffusion}

Unlike SCM-based approaches that explicitly enforce causal relationships through predefined graphs, our method operates \textit{without} imposing assumptions about attribute dependencies. We leverage \textit{language-guidance} as our conditioning mechanism by converting attribute labels into textual descriptions. We implement LANCE, which enhances Stable Diffusion v1.5~\cite{Rombach_2022_CVPR} with null-text inversion to overcome identity-preservation limitations in DDIM when generating counterfactual images \cite{prabhu2023lance}. 

\noindent\textbf{Training} To effectively adapt the generation process to medical imaging distributions while preserving semantic understanding capabilities, we selectively fine-tune the denoising U-Net component while maintaining the pre-trained CLIP encoder's capabilities. We structure prompts to include demographic attributes alongside clinical findings. Specifically, for MIMIC-CXR, our dataset of choice, prompts follow the template: \texttt{Chest X-ray of [age]-years old [race] [sex] subject with [finding(s)]}.

% Image synthesis is performed by LANCE~\cite{prabhu2023lance}, which integrates null-text inversion in order to address DDIM’s well-known limitations in identity-preserving generation. To ensure stability while adapting to the medical imaging domain, the CLIP text encoder and VAE encoder/decoder are frozen while fine-tuning only the denoising U-NET component. This approach retains the CLIP encoder’s broad semantic knowledge while aligning image generation with the original (factual) image distribution. 

% When fine-tuning stable diffusion, the weights of the U-net are modified during the denoising diffusion process (see Figure~\ref{fig:architecture}. We convert MIMIC-CXR's binary disease labels into text captions with a fixed template. 
% We further extend the prompts to include demographic attributes (age, sex, and race) alongside disease and device labels. For example, in the case of the chest Xray data, the text prompts for training are structured as follows: \texttt{Chest X-ray of [age]-years old [race] [sex] subject with [finding(s)]}.

\noindent\textbf{Inference} During inference, generating counterfactual medical images with the fine-tuned model is easily accomplished through a slight modification of the text prompt in a zero-shot manner. In line with interventions in Structural Causal Models (SCMs), counterfactual image generation only requires specifying the attributes we explicitly wish to change. For example, if we want to change the age, the original prompt, \texttt{Chest X-ray of [age]-year old subject with [non-findings]}, is modified to \texttt{Chest X-ray of [new age]-year old subject}, without explicitly defining what disease attribute should be. This approach is distinct from one-hot conditioning, where attributes must be fully defined. Instead, we allow the model to generate the "missing" attributes, based on the specified changes and the image embedding, which provides context and aids the model in inferring the other relevant attributes.

% During inference, generating counterfactual medical images with the fine-tuned model is easily accomplished through a slight modification of the text prompt in a zero-shot manner.
% For instance, performing an intervention to change a patient's sex from male to female, the original prompt  \texttt{Chest X-ray of [age]-year old [\underline{male}] subject} is modified to \texttt{Chest X-ray of [age]-year old [\underline{female}] subject}, resulting in the generated counnterfactual image  As such, the image synthesis  is designed to perform classifier-free guidance.
%To ensure the quality of our counterfactuals, we employ evaluation metrics from \cite{kumar2025PRISM}. Specifically, we measure how well the generated image aligns with the intended text modifications using an editing score $S_\text{CLIP}$ (Eq.~\ref{eq:edit_score}). Following \cite{prabhu2023lance}, we combine this with directional similarity measures \cite{gal2022stylegan} and filter out generations where $S_\text{CLIP}<0.1$ to maintain high-quality outputs.
%\begin{equation}\label{eq:edit_score}
    %S_{\text {CLIP}}=\frac{\Delta X \cdot \Delta T}{\|\Delta X\|\|\Delta T\|}, \quad \text { where } \quad \begin{aligned}
%\Delta X & =E_I\left(X'\right)-E_I\left(X\right), \text { and } \\
%\Delta T & =E_T\left(T'\right)-E_T\left(T\right) 
%\end{aligned}
%\end{equation}
\subsection{Evaluation of Synthesized Counterfactuals}
% Validating the synthesized counterfactual (CF) images is inherently challenging due to the absence of ground truth. 
% % Additionally, we leverage domain knowledge from the literature to qualitatively validate the generated images.

\noindent\textbf{Qualitative Evaluation}: We aim to validate the effectiveness of our method based on several key criteria: (i) \textbf{Precision in High-Fidelity Counterfactual Image Editing}: Our approach enables the generation of high-resolution images with precise modifications to the targeted attribute. This includes accurately removing complex medical devices (e.g., pacemakers, tubes) that vary significantly across the dataset using a single prompt: "Chest X-ray of a subject without support devices."
(ii) \textbf{Comparison with SOTA SCM methods}: We evaluate the effectiveness of our fine-tuned Stable Diffusion model in generating CF images, we compare its outputs to those produced by a SOTA baseline method that explicitly incorporates SCMs~\cite{ribeiro2023high}. This comparison aims to assess whether our model—despite not enforcing causal relationships through explicit constraints—can generate CF images that are both visually plausible and consistent with expected causal dependencies.
% We evaluate our method against a leading SCM model in medical imaging~\cite{ribeiro2023high} that employs a hierarchical variational autoencoder (HVAE). This comparison assesses how well our approach can achieve the same targeted modifications as the SCM method without relying on explicitly defined causal models.
(iii) \textbf{Unveiling Hidden Data Patterns}:  By generating targeted counterfactual modifications, our method uncovers latent correlational or causal relationships that are not explicitly annotated in the dataset and cannot be identified through conventional statistical analysis.
%To evaluate the effectiveness of our fine-tuned Stable Diffusion model in generating CF images, we compare its outputs to those produced by a SOTA baseline method that explicitly incorporates Structural Causal Models (SCMs)~\cite{ribeiro2023high}. This comparison aims to assess whether our model—despite not enforcing causal relationships through explicit constraints—can generate CF images that are both visually plausible and consistent with expected causal dependencies. Additionally, we leverage domain knowledge from the literature to qualitatively validate the generated images. Disease, specifically pleural effusion, is characterized by rounding of the costophrenic angle, augmented lung opacity, and reduced clarity of the diaphragm and lung fissures~\cite{leuallen1955pleural,gayen2022malignant}. For sex, a distinguishable feature is the presence of breast tissue, where female chest radiographs typically exhibit increased soft tissue density over the lower chest wall~\cite{rajee2021gender}. In contrast, male radiographs tend to show more uniform density without prominent breast shadows~\cite{xue2018using,adleberg2022predicting,paul2022demographic}.

\noindent\textbf{Quantitative Evaluations:} For quantitative evaluation, the following metrics ensure that CF images meet key criteria: (i) \textbf{Perceptual Similarity}: To assess whether the CFs maintain visual consistency with the original image, we use the \textit{Learned Perceptual Image Patch Similarity (LPIPS)} score~\cite{zhang2018unreasonable}. This score measures the weighted $\ell_2$ distance across deep feature representations from multiple neural network layers (we use AlexNet~\cite{krizhevsky2012imagenet}). Lower LPIPS scores indicate greater visual similarity, suggesting better anatomical consistency between the original and counterfactual images. with AlexNet~\cite{krizhevsky2012imagenet} architecture. %For a reference image $I_{\text{orig}}$ and its CF version $I_{\text{CF}}$, the LPIPS score measures the weighted $\ell_2$ distance across deep feature representations from multiple layers of a neural network:
%    \[
%    d(I_\text{orig}, I_\text{CF}) = \sum_{l} \frac{1}{H_l W_l} \sum_{h,w} \left\| w_l \odot (\hat{y}^{l}_{hw} - \hat{y}^{l'}_{hw}) \right\|^2_2,
%    \]
%    where $\hat{y}^{l}$ and $\hat{y}^{l'}$ are the normalized feature activations from layer $l$, and $w_l$ scales the activations channel-wise. 
(ii) \textbf{Identity Preservation}: To quantify whether the subject's identity is preserved after the intervention, we compute the \textit{L1 distance}~\cite{mothilal2020explaining,nemirovsky2020countergan} between the factual and counterfactual images. A lower L1 distance suggests better identity preservation, indicating minimal deviation from the original image in pixel space.
(iii)\textbf{Effectiveness}: To measure if each intervention was successful, we leverage pre-trained classifiers \( C_\mathcal{A} \)to measure the \textit{counterfactual prediction gain (CPG)}~\cite{nemirovsky2020countergan,kumar2023debiasing} for attribute \(\mathcal{A} \) of interest. The CPG quantifies the change across the classifier's decision boundary, ensuring that counterfactual interventions effectively lead to the desired attribute modification - $
\text{CPG($\mathcal{A}$)} = \mathbb{E}| C_\mathcal{A}(I_{\text{CF}}) - C_\mathcal{A}(I_{\text{orig}}) |\ \forall \mathcal{A}\in\text{\{disease, sex, race\}}$. A higher CPG value indicates that the counterfactual intervention has successfully altered the desired attribute.

%% file: sec/3_experiments.tex
\section{Experiments \& Results}

\subsection{Dataset and Implementation Details}
We perform experiments on a publicly available dataset, MIMIC-CXR~\cite{johnson2019mimic}, with a training/ validation/ test split of 70/ 15/ 15, see Table~\ref{table:data-distribution}. The dataset has extensive metadata for different races. For the fine-tuning, the \texttt{findings} refer to all 14 diseases available in the metadata including support devices, \texttt{sex} takes the value male/female, \texttt{race} can be   %\footnote{For simplicity, we consider the columns (in the training, validation and test split) `BLACK/AFRICAN AMERICAN', `BLACK/CAPE VERDEAN', `BLACK/CARRIBEAN ISLAND', `BLACK/AFRICAN' as `BLACK' and the columns `ASIAN - CHINESE', `ASIAN', `ASIAN - SOUTH EAST ASIAN', `ASIAN - ASIAN INDIAN', `ASIAN - KOREAN' as `ASIAN'.}. 

\begin{table}[t]
\centering
\caption{Distribution of patients across different data splits. M:Male, F:Female, PE: Pleural Effusion, A:Asian, W:White, and B:Black}

\begin{tabular}{ccccc}
\hline
    & \begin{tabular}[c]{@{}l@{}}Sex [\%]\\ {[}M/F{]}\end{tabular} & \begin{tabular}[c]{@{}l@{}}PE [\%]\end{tabular} & \begin{tabular}[c]{@{}c@{}}Race [\%]\\ {[}A/W/B{]}\end{tabular}& \begin{tabular}[c]{@{}l@{}}[\#]\\ Samples \end{tabular} \\ \hline \hline
Train      & 53.8 / 46.2    &     20.5  & 2.9 / 54.9 / 14.3   & 170333 \\
Validation &  54.0 / 46.0   &     20.5 & 3.1 / 54.9 / 14.4   & 36500 \\
Test       &  53.9 / 46.1   &     20.5  & 2.9 / 54.8/ 14.3   & 36501 \\ \hline
\end{tabular}

\label{table:data-distribution}
\end{table}

We use the DDIM scheduler with null text inversions similar to~\cite{prabhu2023lance} for generating CFs images during inference. %The hyper-parameter details, along with the code, are available in the GitHub repository\footnote{code available on GitHub post acceptance}. % We compare \textcolor[HTML]{4D9900}{our method} with the state-of-the art method, \textcolor[HTML]{FF33FF}{baseline}, for generating high-fidelity counterfactual images based off a structural causal mode. 
Note that our method synthesizes images at resolution $512\times512$ while the baseline method is implemented at a resolution $192\times192$.
\subsection{Results}
\subsubsection{Qualitative Evaluations} 
\begin{figure}[t]
    \centering
    \includegraphics[width=\textwidth]{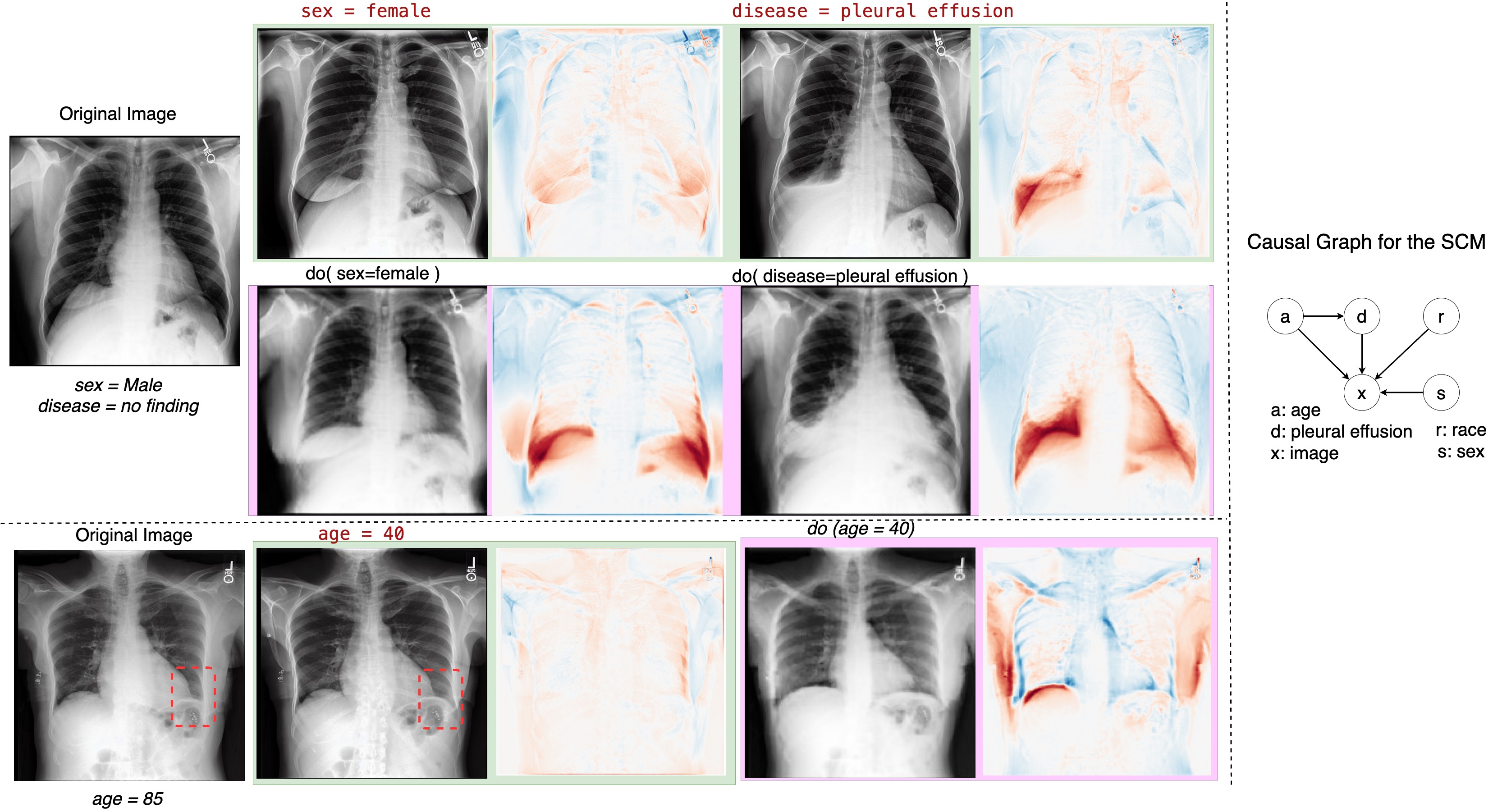}
    \caption{Comparison of counterfactual image generation results using [Row 1] \textcolor[HTML]{4D9900}{proposed method} vs. [Row 2] \textcolor[HTML]{FF33FF}{baseline}, a SOTA method that employs an explicit SCM for generation. For the proposed method, counterfactual images are generated by modifying the text prompt against those generated by performing interventions using the do(.) operator in the baseline method for the attributes: sex and pleural effusion. [Row 3]: Modifications on the age attribute. According to the \textcolor[HTML]{FF33FF}{baseline} pre-defined SCM, there exists an edge between age (a in the SCM) and pleural effusion (d), indicating a possible causal effect. Without defining the explicit SCM, \textcolor[HTML]{4D9900}{proposed method} demonstrates an effect on pleural effusion when modifying age. The regions anticipated to undergo changes are indicated in \textcolor{red}{boxes}. Note that the SCM method significantly alters the synthesis of CF images, resulting in changes to the subject's anatomical structure.}
    \label{fig:compare_scm}
\end{figure}

Our high-resolution results not only align with the baseline method to what interventions in the image should be made, but also maintain high fidelity to the original image \ref{fig:compare_scm}. The baseline method’s pre-defined SCM includes an edge between age and pleural effusion, suggesting a potential causal relationship. Interestingly, without the need for an explicit SCM, the fine-tuned VLM also demonstrates an effect on pleural effusion when age is modified, Figure \ref{fig:compare_scm} Row 3. To further validate the viability of these generated counterfactual images, we consulted an expert radiologist for assessment.

Notably, the fine-tuned vision-language model associates cardiomegaly specifically with pacemakers -- although only the general term support devices is given in the data -- when applying prompt-based modifications. This is evident when prompting the model to remove cardiomegaly, as it also removes the pacemaker. Cardiomegaly, a condition involving heart enlargement~\cite{amin2022cardiomegaly,alghamdi2020study}, is known to be associated with pacemaker implantation~\cite{fung2007pacing}, while pacemakers themselves may contribute to cardiomegaly~\cite{koo2017pacing,gul2024pacemaker,khan2023case}. This bidirectional relationship introduces a potential spurious correlation in the dataset. From a causal standpoint, removing cardiomegaly should not necessarily lead to pacemaker removal, as other medical conditions—undetectable in radiology images—can also necessitate pacemaker implantation~\cite{mesquita2024biological,wood2002cardiac}. This highlights the ability of fine-tuned vision-language models to uncover hidden relationships due to their generative capacity. However, this also underscores a critical caution: while these models enable precise image editing, their changes may reflect spurious correlations rather than faithful causal relationships, reinforcing the need for careful consideration of how counterfactual generation is performed.  
\subsubsection{Quantitative Evaluations}
The quantitative results for the metrics discussed are summarized in Table~\ref{tab:quant}. Both the approaches achieve similar LPIPS and L1 scores, suggesting that the CF images synthesized using both the methods are more visually similar and preserve subject identity effectively. Additionally, the higher CPG score indicates that our approach achieves the expected attribute changes (sex, disease, race, and age) more successfully, validating the consistency of the synthesized counterfactuals.

\begin{figure}
    \centering
    \includegraphics[width=\linewidth]{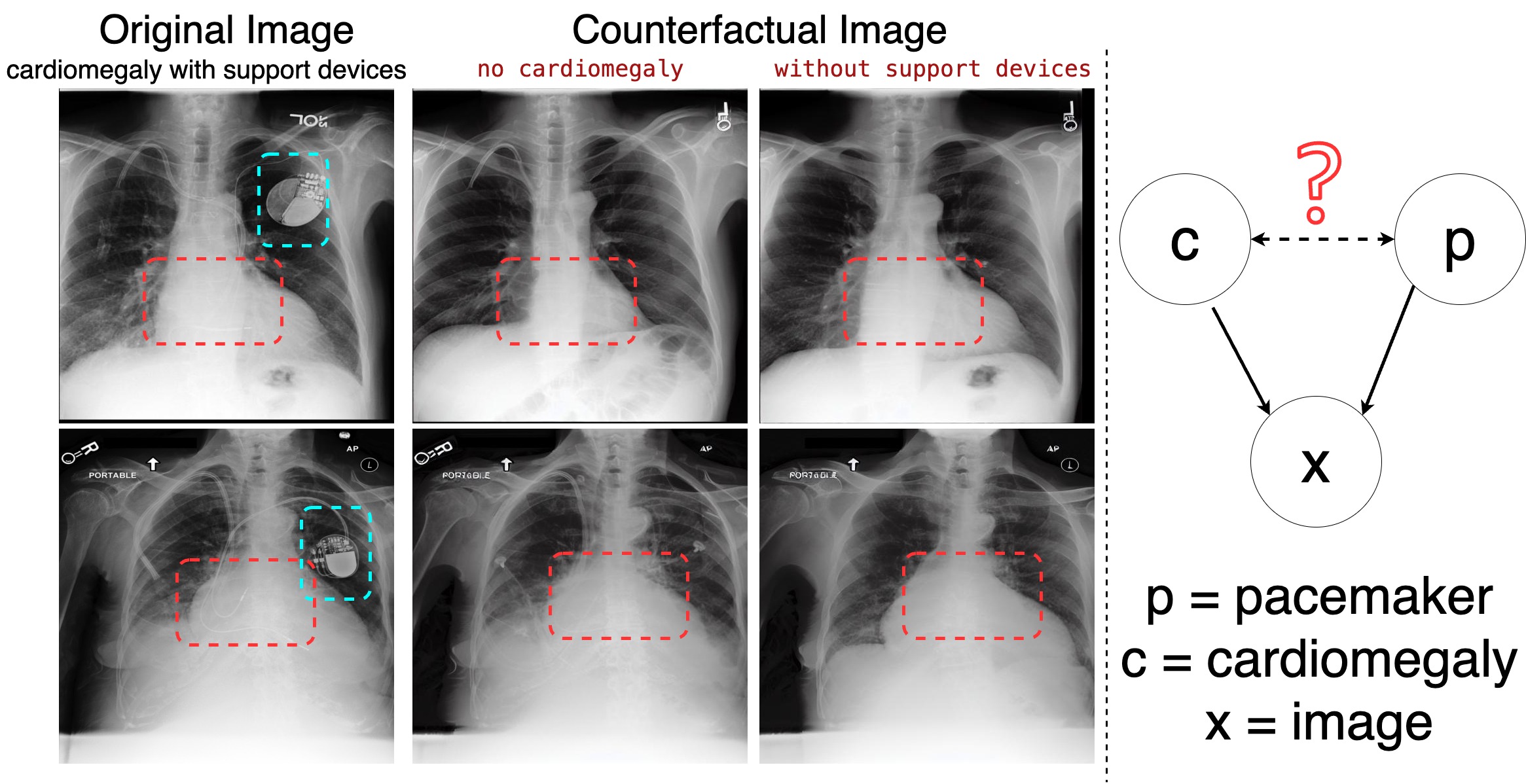}
    \caption{Revealing hidden image-attribute relationships from prompt modifications. [Column 1]: Original Images with corresponding prompts \texttt{Chest X-ray with cardiomegaly and support devices}. [Column 2]: Counterfactual Image with modified text prompt \texttt{Chest X-ray with no cardiomegaly}. [Column 3:] Counterfactual image with modified text prompt \texttt{Chest X-ray with no support devices}. Notably, removing \textcolor{red}{cardiomegaly} also results in the specific removal of the \textcolor{cyan}{pacemaker}, but not the other support devices, suggesting a hidden correlation in the training data. This relationship is supported by the literature~\cite{koo2017pacing,gul2024pacemaker,khan2023case}. (Differences are best seen when zoomed in.)}
    \label{fig:causation}
\end{figure}
\begin{table}
\centering
\caption{Comparison of LPIPS, L1 and CPG on the test set.}

\begin{tabular}{cc|c|c|c}
\hline
\multicolumn{1}{l}{} &  & do(sex) & do(disease) & do(race) \\ \hline
\multirow{2}{*}{\textbf{LPIPS}$\downarrow$ } &  Baseline & 0.15 & 0.17 & 0.10 \\
 & Ours & \textbf{0.11} & \textbf{0.12} & 0.10 \\ \hline
\multirow{2}{*}{\textbf{L1}$\downarrow$ } & Baseline & 16.62 & 16.65 & \textbf{8.92} \\
 & Ours & \textbf{15.51} & \textbf{14.21} & 9.05 \\ \hline
\multirow{2}{*}{\textbf{CPG}$\uparrow$ } & Baseline & 0.73 & 0.81 & 0.70 \\
 & Ours & \textbf{0.81} & \textbf{0.89} & \textbf{0.75} \\ \hline
\end{tabular}

\label{tab:quant}
\end{table}
\section{Conclusion}
Our work highlights the potential of fine-tuned vision-language foundation models in identifying critical data biases and spurious correlations. By leveraging their ability to generate high-resolution, precisely edited images through language prompts, these models reveal hidden data patterns that were previously undetectable. This has significant implications for the development of VLM-based methods in healthcare, where understanding dataset biases is crucial for building robust, trustworthy, and clinically deployable models. Our findings also underscore the limitations of fine-tuned VLMs, including their susceptibility to spurious correlations. Future work will focus on integrating causal reasoning to mitigate the reliance on biases and enhance the reliability of these models in clinical applications.